\documentclass[11pt,a4paper]{article}
\usepackage[hyperref]{acl2020}
\usepackage{times}
\usepackage{amsfonts}
\usepackage{svg}
\usepackage{graphicx}
\usepackage{tabularx}
\usepackage{latexsym}
\usepackage[]{todonotes}

\usepackage{ wasysym }

\usepackage{microtype}

\aclfinalcopy 
\def\aclpaperid{1917} 

\title{DoQA - Accessing Domain-Specific FAQs via Conversational QA}

\author{Jon Ander Campos\textsuperscript{\rm 1}, Arantxa  Otegi\textsuperscript{\rm 1}, Aitor Soroa\textsuperscript{\rm 1},  
\\ \textbf{Jan Deriu\textsuperscript{\rm 2}, Mark Cieliebak\textsuperscript{\rm 2}, Eneko Agirre\textsuperscript{\rm 1}}\\ 
\textsuperscript{\rm 1}University of the Basque Country (UPV/EHU)\\
\textsuperscript{\rm 2}Zurich University of Applied Sciences (ZHAW)\\
\textsuperscript{\rm 1} \texttt{\{jonander.campos, arantza.otegi, e.agirre, a.soroa\}@ehu.eus} \\
\textsuperscript{\rm 2}\texttt{\{jan.deriu, mark.cieliebak\}@zhaw.ch} }

\date{}

\begin{document}
\maketitle
\begin{abstract}
  The goal of this work is to build conversational Question Answering (QA) interfaces for the large body of domain-specific information available in FAQ sites. 
  We present DoQA, a dataset with 2,437 dialogues and 10,917 QA pairs. The dialogues are collected from three Stack Exchange sites 
  using the Wizard of Oz method with crowdsourcing.  
  Compared to previous work, DoQA comprises well-defined information needs, leading to more coherent and natural conversations with less factoid questions and is multi-domain.
  In addition, we introduce a more realistic information retrieval (IR) scenario where the system needs to find the answer in any of the FAQ documents.
  The results of an existing, strong, system show that, thanks to transfer learning from a Wikipedia QA dataset and fine tuning on a single FAQ domain, it is possible to build high quality conversational QA systems for FAQs  without in-domain training data. The good results carry over into the more challenging IR scenario. In both cases, there is still ample room for improvement, as indicated by the higher human upperbound. 

\end{abstract}

\section{Introduction}
\label{sec:intro}

\begin{figure}[t]
    \centering
    \includegraphics[width=0.45\textwidth]{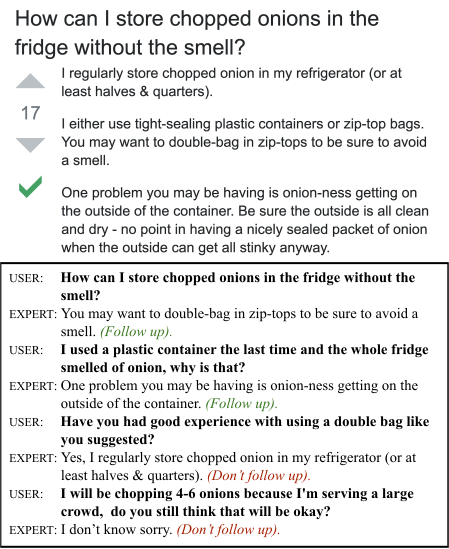}
    \caption{A dialogue about cooking. On top, the original post, comprising a topic and an excerpt of the answer passage. 
    In italics, dialogue acts (cf. Section \ref{sec:datasetcollection}).}
    \label{fig:example_dialog}
\end{figure}

The overarching objective of our work is to access the large body of domain-specific information available in Frequently Asked Question sites (FAQ for short) via conversational Question Answering (QA) systems. In particular, we want to know whether current techniques are able to work with limited training data, and without needing to gather data for each target FAQ domain. 
 In this paper we present \textbf{DoQA}, a task and associated dataset for accessing domain-specific FAQs via conversational QA\footnote{The DoQA dataset is available here: \url{http://ixa.eus/node/12931}}. 
The dataset contains 2,437 information-seeking question/answer dialogues on three different domains (10,917 questions in total). These dialogues are created using the Wizard of Oz technique by crowdworkers that play the following two roles: the \textbf{user} asks questions about a given topic posted in Stack Exchange\footnote{\url{https://stackexchange.com/}}, 
and the \textbf{domain expert} replies to the questions by selecting a short span of text from the long textual reply in the original post. The first question is prompted by the real FAQ question, which sets the topic of interest driving the user questions. In addition to the extractive span, we also allow experts to rephrase it, in order to provide an abstractive, more natural, answer. The dataset covers unanswerable questions and some relevant dialogue acts. We focused on three different domains: Cooking, 
Travel and 
Movies. 
These forums are some of the most active ones and contain knowledge of general interest, making it easily accessible for crowdworkers. DoQA contains two scenarios: in the standard scenario the test data comprises the questions and the target document from which the answers need to be extracted; in the information retrieval (IR) scenario the test data contains the questions, but the target document is unknown, and the system needs to select the documents which contain the answers among all documents in the collection. 



Previous work on conversational QA datasets include CoQA \cite{reddy2018coqa} and QuAC \cite{choi2018quac}. The main focus of CoQA are reading comprehension questions, which are produced with access to the target paragraph. The topic of the questions are delimited by the paragraph, which leads to specific questions about details in the paragraph. \citet{choi2018quac} observed that a large percentage of CoQA answers are named entities or short noun phrases. In QuAC, the topic of the conversation is set by a title and first paragraph of a Wikipedia article about people. The user makes up questions about the person of interest. Note that, contrary to our setting, there is no real information need in any of those datasets, which can lead to less coherent conversations: any question about the paragraph or person of interest is valid, respectively. 

DoQA makes the following \textbf{contributions}. Firstly, contrary to made-up reading comprehension tasks, DoQA reflects real user needs, as defined by a topic in an existing FAQ. Good results on DoQA are of practical interest, as they would show that effective conversational QA interfaces to FAQs can be built.  Secondly, for the same reason, the conversations in DoQA are more coherent, natural and contain less factoids than other datasets, as shown by our analysis. Thirdly, the IR scenario and the multiple domains make DoQA more challenging and realistic. Table \ref{tab:characteristics} summarizes the characteristics of DoQA. 


Although one could question the small size of our dataset, our goal is to test whether current techniques are able to work with limited training data, and without needing to gather data for each target FAQ domain. We thus present results of an existing strong conversational QA model with limited and out-of-domain data. The system trained on Wikipedia data (QuAC) provides some weak results which are improved when fine-tuning on the FAQ dataset. Our empirical contribution is to show that a relatively low amount of training in one FAQ dataset (1000 dialogues on Cooking) is sufficient for strong results on Cooking (comparable to those obtained in the QuAC dataset with larger amounts of training data), but also on two other totally different domains with no in-domain training data (Movies and Travel). In all cases scores over 50 F1 are reported. Regarding the IR scenario, an IR module complements the conversational system, with a relatively modest drop in performance. 
The gap with respect to human performance is over 30 points, showing that there is still ample room for system improvement. 


\begin{table}[t]
\centering
\small
\resizebox{0.48\textwidth}{!}{
\begin{tabular}{lccc}
\hline
                                                     & DoQA  & QuAC   & CoQA    \\ \hline
Real information need        &        \UParrow & &    \\ 
Naturalness        & \UParrow &  &   \\
Dialogue coherence        & \UParrow &  &   \\
Non-factoid questions& \UParrow &  &   \\
Unanswerable questions    &            \UParrow & \UParrow &    \\ 
Dialogue acts&            \UParrow & \UParrow &   \\ 
Multi-domain&            \UParrow &  &  \UParrow  \\ 
IR scenario&            \UParrow &  &   \\ 
\hline
\end{tabular}}
\caption{Summary of the characteristics of DoQA compared to QuAC and CoQA. \UParrow ~for positive. } 
\label{tab:characteristics}
\end{table}

\section{Related Work}

Conversational QA systems stem from the body of work on Reading Comprehension, whose goal is to test the capacity of a system to understand a
document by answering any question posed over its content. Recent work on
the field has resulted in the creation of multiple datasets
\cite{rajpurkar2016squad,trischler2017newsqa,nguyen2016ms,kocisky2018narrativeqa,dunn2017searchqa}.
These datasets are typically composed of multiple question/answer pairs, often
along with a reference passage from which the answer is curated. Whereas the
questions are always in free text form, some datasets represent the answers
as a contiguous span in the reference passage, while others contain free form answers. The former are usually referred as \emph{extractive}, whereas the latter are called \emph{abstractive}. All in all, in these QA datasets the queries are unrelated to each other, and thus there is no dialogue structure involved.

\citet{iyyer2017search} propose to answer complex queries by decomposing them into sequences of single, co-referent queries. The question sequence can be seen as different turns in a dialogue, and each question refers and refines previous ones. The authors present the SequentialQA dataset, which comprises 6K question sequences posed over the content of Wikipedia tables. In the case of our task, it is the user who makes several questions in sequence.

More similar to our work, CoQA~\cite{reddy2018coqa} and
QuAC~\cite{choi2018quac} are two conversational QA datasets
comprising QA dialogues that fulfill the information need of a
user by answering questions about different topics. Similarly to our, both
datasets are built by crowdsourcing, where one person (the questioner) is
presented with a topic and has to pose free-form questions about it. Another
person (the answerer) has to select an answer to the question by choosing an
excerpt from the relevant passage describing the topic. Some of the
questions in both datasets are unanswerable, and access to previous questions and answers are 
needed in order to answer some of the questions.

CoQA contains 127k questions with answers, obtained from 8k conversations about passages from broad domains, ranging from children stories to science. The answers are also excerpts from the relevant passage, but answerers have the choice of reformulating them. The authors report that $78\%$ of the answers had at least one edit. Although reformulating answers can yield to more natural dialogues, \citet{yatskar2018qualitative} showed that span based systems can in principle obtain a performance up to $97.8$ points F1, showing that editing the answers does not yield to systems with better quality. In CoQA, both questioner and answerer have access to the full passage, which guides the conversation towards the specific information conveyed in it. 


QuAC is a dataset that contains 14k information-seeking question answering dialogues. The dialogues in QuAC are about a specific section in Wikipedia articles about people. The answerer has access to the full section text, whereas the questioner only sees the section's title and the first paragraph of the main article, which serves as inspiration when formulating the queries. QuAC also contains dialogue acts in each turn, which are useful when collecting the dialogues, as they can be used by the answerer to indicate to questioner whether to continue making questions about the last answer or drift to other aspects of the topic.  
We will compare CoQA and QuAC in more detail in Section \ref{sec:analysis}.

Previous conversational QA  datasets provide the relevant document or passage that contain the answer of a query. However, in many real world scenarios such as FAQs, the answers need to be searched over the whole document collection. In related question answering research, \citet{chen2017drqa} and \citet{watanabeds2017questionanswering} combine retrieval and answer extraction on a large set of documents. In \citep{talmor2018web} the authors propose decomposing complex questions into a sequence of simple questions, and using search engines to answer those single questions, from which the final answer is computed.
We find that requiring the system to search for relevant documents and passages is more realistic, and DoQA is the first conversational QA task incorporating this scenario. 

In contemporary work, \citet{castelli2019techqa} present a question  answering  dataset  for  the  technical support domain which focuses on actual questions posed by users and has a real-world size with only 600 training instances. It also requires systems to examine 50 documents per query. Our work has similar motivations for setting up more realistic tasks, and is complementary in the sense that we cover non-technical domains and conversatioal QA. 

Community Question Answering has been also the focus of two related tasks \cite{nakov2016semeval,nakov2017semeval}, where, given a new question and a collection of pre-existing questions and answers, the systems need to rank the answers that are most useful for answering the new question.

\section{Dataset Collection}
\label{sec:datasetcollection}

This section describes our conversational QA dataset collection process which consists of an interactive task designed for two crowdworkers in Amazon Mechanical Turk (AMT).

\subsection{FAQ Post Selection}
\label{ssec:cookingthreadselection}

We collected topic-answer pairs for the three different domains from the Stack Exchange data dumps. We focused on the Cooking\footnote{\url{https://cooking.stackexchange.com/}},
Travel\footnote{\url{https://travel.stackexchange.com/}} and 
Movies\footnote{\url{https://movies.stackexchange.com/}} domains, as they are active forums and contain knowledge of general interest, making it easily accessible and attractive for crowdworkers. Note that the posts in Stack Exchange (as in most FAQ sites) comprise broad questions which often require lengthy answers. We refer to the question in the post as \emph{topic}  and to the long answer in the post as \emph{passage} (not to be confused with the actual questions/answers in the collected dialogues). Figure \ref{fig:example_dialog} shows an example of a topic and its corresponding passage for the Cooking domain. More details on post filtering and selection can be found in Appendix A.

\subsection{Crowdsourcing Task}
\label{ssec:interactivetask}

For the annotation process, we defined a HIT in AMT as the task of generating a dialogue about a specific topic between two workers (the specifications of the defined HIT can be found in Appendix B). 
One of the workers (the user) asks questions to the second one (the domain expert) about a certain topic from a Stack Exchange Cooking, Travel or Movies thread. The worker who adopts the user role has access to a small paragraph that introduces the topic. Having this information, he must ask free text questions. The first question of every dialogue must be the title of the topic that appears in the title of the Stack Exchange thread. The domain expert has access to the whole answer passage and he/she answers the query by selecting a span of text from it. In order to make the dialogue look more natural, the domain expert has the opportunity to edit the answer, but note that if he does so the answer will not match the content of the text span anymore. Therefore, and following \citet{yatskar2018qualitative}, we motivate minimal modifications by copying the selected text span directly into the answer field in the web application. In addition to the span of text, the expert has to give feedback to the user with one of the following dialogue acts: an affirmation act, which is is required when the question is a Yes/No question (\textit{yes}, \textit{no} or \textit{neither}); an answerability act, which defines if the question has an answer or not (\textit{answerable} or \textit{no answer}). When no answer is selected, the returned string is ``I don't know''; and a continuation dialogue act, which is used for leading the user to the most interesting topics (\textit{follow up} or \textit{don't follow up}). The last dialogue act is used to minimally guide the user in his/her questions, where the expert can encourage (or dicourage) the user to continue with questions related to his last questions using \textit{follow up} (or alternatively \textit{don't follow up}).  These dialogue acts are the same as in QuAC, but we discarded the \textit{maybe follow up} act from the continuation act because we felt it was not intuitive enough. 

Dialogues are ended when a maximum of 8 question and answer pairs is reached, when 3 unanswerable questions have been asked, or when 10 minutes time limit is reached. The purpose of these limits is to avoid long and repetitive dialogues, because real threads of the selected domains are very focused on a certain topic. 
Dialogues are only accepted if they have a minimum length of 2 question and answer pairs and if they have at least one answer that is not ``I don't know sorry''.
    
The data collection interface is based on 
CoCoA\footnote{\url{https://github.com/stanfordnlp/cocoa} \cite{he2017learning}}, which we modified. The interfaces for the user and expert are shown in Appendix C. 

\subsection{Dataset Details}
\label{ssec:datasetdetails}


\begin{table}[t]
\centering
\resizebox{.48\textwidth}{!}{
\begin{tabular}{lrrrrr}
\hline
                     & \multicolumn{3}{c}{Cooking}  & Travel & Movies \\ 
                     & Train & Dev.  & Test  & Test & Test   \\ \hline
Questions            & 4,612 & 911   & 1,797 & 1,713 & 1,884    \\
Dialogues            & 1,037 & 200   & 400   & 400 & 400    \\
Unique sections      & 546   & 162   & 400   & 400 & 400     \\ \hline
Tokens / question            & 10.79 & 10.14 & 10.66 & 10.45 & 9.45    \\
Tokens / answer        & 13.19 & 13.10 & 12.58 & 13.47 & 12.40   \\
Dialogue turns        & 4.47  & 4.55  & 4.49  & 4.28 & 4.71    \\
Extractive \%        & 69.68 & 67.18 & 66.95 & 65.44 & 74.15   \\
Abstractive \%       & 30.32 & 32.82 & 33.05 & 34.56 & 25.85   \\
Yes/No \%            & 20.22 & 21.07 & 22.20 & 25.10 & 18.05   \\
I don't know \%      & 27.55 & 27.33 & 29.71 & 22.83 & 29.41   \\ \hline
\end{tabular}}
\caption{Statistics of the different domains of DoQA. }
\label{tab:train/test/dev}
\end{table}

Following usual practice, we divided the main Cooking dataset into a train, development and test splits. For the other two domains, Travel and Movies, we only have the test split. Statistics for all the domains and splits are shown in Table \ref{tab:train/test/dev}. 

The splits of the Cooking dataset have very similar characteristics, so we can expect them to be valid representatives of the whole Cooking dataset. In the test splits we do not allow more than one dialogue about the same section, as it can end up producing inaccurate evaluation of the models.

\subsection{Collecting Multiple Answers}
\label{ssec:multipleanswers}

In order to estimate the performance of a human in the task, we collected additional answers for the test splits for the three domains in a second round, after having completed the dialogues. 
For each question in the dialogues collected in the first round, we show to the worker the previous questions and answers in the dialogue (if available), and he has to provide an answer span. The interface for the collection of multiple answers can be seen in Appendix D.

\subsection{Information Retrieval Scenario}
\label{ssec:ir}

In the usual setting for this kind of tasks, the system is given the question and the passage where the answer is to be extracted from. In a realistic scenario, however, relevant answer passages that may contain the answer will need to be retrieved first. More specifically, if a user has an information need and asks a question to a conversational QA system on a FAQ, the system can search for similar questions which have already been answered, or the system can directly search in existing answer passages. In other words, there are two ways to check automatically if the forum contains a relevant answer passage to a new question: (1) question retrieval, where relevant or similar questions are searched (and thus, the answer for this relevant question is taken as a relevant answer), and (2) answer retrieval, where relevant answers are searched directly among existing answers.

We added information about both relevant cases to the main Cooking dataset, in the form of the 20 most relevant answer passages for each dialogue in the dataset. We followed a basic approach to get these relevant answer passages. We created two separate indexes using an IR system\footnote{Solr \url{https://lucene.apache.org/solr/}} for the two mentioned approaches, question and answer retrieval. For the former, we indexed the original topics posted in the forum; and for the latter, we indexed the answer passages for each post in the forum. Then, for each dialogue in the development and test splits, the top 20 documents were retrieved using the first question of the dialogue. Given that the dialogues are about a single topic, we only use the first question in the dialogue, and then use the retrieved passages for the rest of questions in the dialogue as well.  


The question retrieval approach yields very good results (0.94 precision at one), as expected, as  the crowdworker doing the questions  has access to the topic when asking the first question and usually did minor edits. 
The results for answer retrieval are more modest, 0.54 precision at one.
The results section shows the results of the conversational QA system when relying on the passages returned by the IR module.

\section{Dataset Analysis}
\label{sec:analysis}

\paragraph{Overall statistics}

\begin{table}[t]
\centering
\small
\resizebox{0.48\textwidth}{!}{
\begin{tabular}{lrrr}
\hline
                         & DoQA  & QuAC   & CoQA    \\ \hline
Questions                & 10,917          & 98,407          & 127,000 \\ 
Dialogues                & 2,437          & 13,594 & 8,399            \\ \hline
Tokens / question        & 10.43 & 6.5             & 5.5              \\ 
Tokens / answer          & 12.99          & 14.6   & 2.7              \\ 
Dialogue turns     & 4.48           & 7.2             & 15.2   \\ 
Extractive \%            & 69.13          & 100    & 66.8             \\ 
Abstractive \%           & 30.87          & -               & 33.2    \\ 
Yes/No \%                & 21.01          & 25.8  & -                \\
I don't know \%          & 27.47 & 20.3            & 1.3              \\ \hline
\end{tabular}}
\caption{Statistics of DoQA compared to QuAC and CoQA.}
\label{tab:stats}
\end{table}

In this section we present an quantitative and qualitative analysis of DoQA and we compare them to similar conversational datasets like QuAC and CoQA, stressing its similarities and differences.

Table \ref{tab:stats} shows the overall statistics of DoQA, together with the statistics of QuAC and CoQA. As can be seen, DoQA has the smallest amount of questions and dialogues. However, other features makes it very interesting for the research of conversational QA. For instance, the average tokens per questions and answers ($10.43$ and $12.99$, respectively) are closer to real dialogues if we compare to the other datasets. Specially CoQA has very short questions and answers on average, suggesting that CoQA is closer to factoid QA than dialogue, as human dialogues tend to be longer and convoluted, not just short answers. DoQA has the lower ratio of questions per dialogue, which is expected, as most of the dialogues are about a very specific topic and the user is satisfied and gets the answer without the need of long dialogues. CoQA ends up on having almost all of its questions answerable, facing the same issues as SQuAD 1.0 \citep{rajpurkar2016squad} that motivated the addition of unanswerable questions in SQuAD 2.0 \citep{rajpurkar2018know}.

We also have the results of a short survey that workers had to respond to at the end of each HIT. On the one hand, the user had to give feedback on how satisfied was with the answers of the expert in a scale of 1-5. The average satisfaction was 3.9. On the other hand, the expert had to give feedback on how sensible were the questions and the helpfulness of the answers. The average scores obtained were 4.27 and 4.10, respectively, which makes the AMT task satisfactory.

\paragraph{Naturalness}

One of the main positive aspects of our dataset is the naturalness of the dialogues that other similar datasets like QuAC do not have. The answers of DoQA come from a forum where the answer text is directed to a person who posted the question, and does not come from a much formal text like Wikipedia, as it is the case of QuAC. The naturalness and casual register of the former it is more adequate than the formal register of the latter for a conversational QA system. The dialogue in Figure \ref{fig:example_dialog} is a clear example of such naturalness, where the expert answers to the user with casual and directed expressions like ``\textit{You may want}'' and ``\textit{you may be having}''. To verify whether dialogues in DoQA are more natural than the ones in QuAC, we sample randomly 50 dialogues in DoQA Cooking domain and QuAC and performed A/B testing to determine which of the two dialogues is more natural. This test showed that 84\% of the times DoQA dialogues are more natural. 

This naturalness is probably caused because a dialogue in DoQA is started by a user with a very specific aim or topic to solve in mind, and thus, follow-up questions are very related to previous answers, and all the questions are set within a context. In contrast, dialogues in QuAC do not show so clear objective and questions seem to be asked randomly. Dialogues in DoQA are ended when the initial information need of the user is satisfied and this adds naturalness to dialogues. 

Further analysis of the samples showed that answers in DoQA seem to be more spontaneous because they have more orality aspects, such as higher level of expressivity (``\textit{Normally when I try they end up burned not crispy!}'', ``\textit{My biggest worry here would be...}'', ``\textit{hey let's not be hasty}''), opinions (``\textit{I came across a suggestion to cover the lid...}'', ``\textit{I'd recommend simply adding...}'', ``\textit{It sounds like fermentation to me}'') and humor (``\textit{well yeah but booze is booze}''). Contrarily, answers in QuAC are more hermetic and do not show any features of orality or spontaneity that a dialogue should have. All these features make DoQA dialogues look more natural. 

We also analyzed the remaining 16\% cases where DoQA dialogues appear less natural. In most of these dialogues there were responses that did not really answer the question. The following question (Q) and answer (A) pairs are good examples of it: (Q) ``\textit{Is the taste going to be significantly different?}'' (A) ``\textit{there is cornstarch in confectioner's sugar}''; (Q) ``\textit{how about reheating?}'' (A) ``\textit{When you defrost it, do so in your fridge leaving it overnight so that it defrosts gradually}''; (Q) ``\textit{Can I use my potatoes or carrots if they already have some roots?}'' (A)	``\textit{The green portions of a potato are toxic}''. In some of these cases the correct answer for the respective question is not in the answer text provided to the expert. If this was the case, the expert should answer "I don't know", instead of giving a nonsense answer.

\begin{table*}[t]
\centering
\resizebox{\textwidth}{!}{
\begin{tabular}{lrll}
\hline
\multicolumn{2}{l}{Bigram prefix}   & \%        & Example             \\ \hline
What &is                & 30.8      & What is the purpose of adding water to an egg wash?  \\
(16.6\%) &are                                & 8.0       & What are other methods to sharpen a knife?        \\ \hline
How &  do                               & 24.0      & How do you properly defrost frozen fish?     \\ 
(15.1\%) &  long                             & 21.9      & How long should I cook it in the microwave?   \\ \hline
Is &  there                            & 52.8      & Is there a special tool available for cracking open a pistachio?   \\ 
(10.5\%) &  it                               & 19.8      & Is it safe to cook with rainwater? \\ \hline
Do &  you                              & 70.7      & Do you have any advice for storing green onions?  \\
(7.6\%)&  I                                & 16.1      & Do I have to peel the apples?  \\ \hline
Can &  I                                & 52.8      & Can I put them back in the oven to reheat? \\ 
(5.5\%)&  you                              & 25.3      & Can you explain the science behind this cooking procedure?\\ \hline
I &  have                             & 19.6      & I have been told that frying it would make it tastier, but is it healthier to grill or fry?                \\ 
(5.0\%)&  am                               & 15.3      & I am cooking for somebody who doesn't eat shellfish, so is the fish sauce safe?            \\ \hline
Why &  is                               & 22.1      & Why is it important to increase the fermentation time?    \\ 
(3.5\%) &  does                             & 21.7      & Why does my custard pudding taste like raw eggs?   \\ \hline
\end{tabular}}
\caption{Most frequent initial words and bigrams in questions (Cooking domain).}
\label{tab:questions}
\end{table*}

\paragraph{Question types} Table \ref{tab:questions} includes the most frequent two initial words of the questions in the Cooking dataset along with their percentages of occurrences and some examples.
Most of the questions start with \textit{what} and \textit{how} ($16.6\%$ and $15.1\%$ of the questions, respectively), which are also the most frequent in QuAC and CoQA. Contrary to them, the questions in the Cooking dataset do not refer to factoids, with the exception of ``How long'' questions. The questions in DoQA require long and complex answers. In contrast to this, in CoQA and QuAC many of the most frequent initial words such as \textit{who}, \textit{where}, and \textit{when} indicate factoid questions. In order to confirm this fact, we manually inspected $50$ random questions from the Cooking domain and QuAC datasets. This analysis revealed that 66\% of the questions are non-factoid in the DoQA Cooking domain, showing that most of the questions are open-ended. These amount is larger than in QuAC, as in our analysis for QuAC we found that only 36\% of the questions are non-factoid. These values differ slightly from those reported by \citet{choi2018quac}, as they say that about half of questions
are non-factoid. 


\paragraph{Context or history dependence} The manual analysis also shows that $61\%$ of the questions are dependent on the conversation history, as many questions have coreferences to previous questions or answers in the dialogue. For example, ``\textit{What are other methods to sharpen a knife?''}, ``\textit{How long should I cook it in the microwave?''}, ``\textit{Can you explain the science behind this cooking procedure?''}.
Moreover, we could note that less than $1\%$ ask further advice or tips about the current topic, confirming that these conversations are about specific topics where the user is satisfied with the expert answers after a few questions.

\paragraph{Dialogue coherence}


Related to the just mentioned fact that the user does not usually ask any other tips, users in DoQA do not tend to switch topics in a dialogue. In order to confirm it, we performed another A/B testing to the same 50 dialogues samples of the DoQA Cooking domain and QuAC to determine which of the two dialogues is more coherent, that is, which dialogue has a smoother flow. This test revealed that in 64\% of the cases dialogues of DoQA are more coherent than QuAC. Only in 10\% of the cases dialogues of DoQA are less coherent, with the remaining 26\% equally coherent. We analyzed the 10\% and saw that they contain similar questions one after the other, or repeated answers in the same dialogue.

\paragraph{Summary} Table \ref{tab:characteristics} summarizes the positive characteristics of DoQA compared to the similar datasets like QuAC and CoQA.


\section{Task Definition}
\label{sec:task}

Given a textual passage and a question, traditional QA systems find an answer to the question within the passage. Conversational QA systems are more complex, as they need to deal with a sequence of possibly inter-dependent questions. That is, the meaning of the current question may depend on the dialogue history. For this reason, a dialogue history comprised by previous question/answer pairs is also provided to the system. In addition, some dialogue acts have to be predicted as an output: yes/no answers, which are required for affirmation questions, and continuation feedback, which might be useful for information-seeking dialogues.

We denote the answer passage as $p$, the dialogue history of questions and respective ground truth answers as $\{q_1,a_1,...q_{k-1},a_{k-1}\}$, current question as $q_k$, the answer span $a_k$ which is delimited by its starting index $i$ and ending index $j$ in the passage $p$, and dialogue act list $v$. The dialogue act list contains \textit{\{yes,no,-\}} values for predicting affirmation and \textit{\{follow-up,don't follow-up\}} for continuation feedback.

\section{Baseline Models}
\label{sec:baseline}

We present two strong baseline models to address our task. Although the state-of-the-art evolves quickly, our choice has the benefit of simplicity and strong performance. 

\paragraph{BERT}
We took the fine-tuning approach for QA of BERT, which predicts the indexes $i$ and $j$ of the $a_k$ answer span given $p$ and $q_k$ as input. This baseline has shown strong performance on QA datasets such as SQuAD~\citep{devlin2018bert}. 

\paragraph{BERT+HAE}
The previous baseline does not model dialogue history. We used BERT with History Answer Embedding (HAE) as proposed by \citet{qu2019} as a baseline that deals with the multi-turn problem, as this is the publicly available system that performs best in the QuAC leaderboard\footnote{accessed on August 20, 2019}. The system introduces dialogue history $\{q_1,a_1,...q_{k-1},a_{k-1}\}$ to BERT by adding a history answer embedding layer, which learns whether a token is part of history or not.

\begin{table*}[t]
\centering
\resizebox{\textwidth}{!}{
\begin{tabular}{ll|ccc|ccc|ccc}
          &          & \multicolumn{3}{c|}{Cooking} & \multicolumn{3}{c|}{Travel} & \multicolumn{3}{c}{Movies} \\ \hline
Setting & \multicolumn{1}{l|}{Model}    & F1      & HEQ-Q   & F1all   & F1    & HEQ-Q    & F1all   & F1    & HEQ-Q    & F1all   \\ \hline
Native & BERT     & 40.1    & 35.1    & 38.3     &    36.2  &  34.8    &    34.8     &   36.1  &     33.5    &   35.0    \\
& BERT+HAE & 47.8    & 43.0    & 45.9    &    44.0   &   37.4       &   42.9       &    42.8   &      37.1    &      41.9    \\ \hline
Zero-shot & BERT     & 40.2    & 34.7    & 38.9     &   34.0   &     30.1    &     33.1    &    38.2   &     33.2     &      37.4    \\
& BERT+HAE & 46.2    & 42.0    & 44.5     &    42.7   &      37.1    &    42.3      &   45.4    &  41.4        &      44.8    \\ \hline
Transfer & BERT     & 43.3   & 37.8    & 42.4   &   40.6    &     33.6     &     40.1     &     41.8  &    36.3      &       41.3   \\
& BERT+HAE & 53.2    & \textbf{48.3}   &  51.4     &   50.8    &    42.1      &     50.6      &    51.6   &      44.3    &    51.5      \\ \hline
Transfer& BERT     &  43.1   &  37.0   & 42.0  &  40.6   &     33.4    &    40.5   &  42.0 &    34.5    &   41.6   \\
all & BERT+HAE &  \textbf{53.4}   &   46.9  &  \textbf{52.7}   &   \textbf{51.6}    &     \textbf{43.3}     &    \textbf{50.9}     &    \textbf{52.1}   &      \textbf{45.2}    &     \textbf{51.7}     \\ \hline
\multicolumn{2}{l|}{Human} &  -  &  100.0 & 86.6  &  - &  100.0 & 87.4  & -  & 100.0 & 88.8\\ \hline
\end{tabular}
}
\caption{Results of the baseline systems in the three DoQA domains (columns) in all four settings (rows). See text for explanation of each row. Note that Travel and Movies results are obtained without any Travel or Movies training data. 
}
\label{tab:transfer-results}
\end{table*}

\section{Evaluation}
\label{sec:evaluation}

\paragraph{Evaluation metrics}
\label{ssec:metrics}
Given the similarity between QuAC and DoQA, we use the same evaluation metrics and criteria used in QuAC. F1 is the main evaluation metric and is computed by the overlap at word level of the prediction and reference answers. As the test set contains multiple answers for each question we take the maximum F1 among them. Note that when computing F1 QuAC filters out answers with low agreement among human annotators. An additional F1-all is provided for the whole set. We also report HEQ-Q (human equivalence score on a question level) which measures the percentage of questions for which system F1 exceeds or matches human F1.

\paragraph{Experimental Setup}
\label{ssec:setup}

We carried out experiments using the extractive information in DoQA, leaving the abstractive information for the future. The parameters we used to train the baseline models are the ones proposed in the original papers. We tested the models in four settings. In the \textbf{native} setting the Cooking DoQA train and dev data are used, the first for training and the second for early stopping. In the \textbf{zero-shot} setting we use QuAC training data for training and early stopping.  In the \textbf{transfer} setting we use QuAC and Cooking DoQA for training. Finally, in the \textbf{transfer all}  setting we additionally use the test data from the other two domains for training.


We also experimented on the IR scenario, using the provided IR rankings (see Section \ref{ssec:ir}), which contain the top $20$ passages for each dialogue. In the first experiment, 
\textit{Top-1}, we just use the top 1 passage and apply the baseline BERT model. In a second experiment,  \textit{Top-20:BERT}, the passages are fed to the BERT model and the passage that contains the answer with highest confidence score is selected. Note that we discard passages that produced ``I don't know'' type of answers. In a third experiment,  \textit{Top-20:BERT*IR}, we select the passage with highest combined score according to BERT and the search engine.

All the reported results have been achieved using the BERT Base Uncased model.

\paragraph{Results}
\label{ssec:results}

Table \ref{tab:transfer-results} summarizes our results. 
In the bottom row we give the human upperbound.
The three metrics used for evaluation behave similarly, so we focus on one (e.g. F1) for easier discussion. We report all three for completion and easier comparison with related datasets.
In all settings and domains the BERT+HAE model yields better results than BERT, showing that \textbf{DoQA is indeed a conversational dataset}, where question and answer history needs to be modelled. 



Regarding the different settings, we first focus on the \textbf{Cooking} dataset. The native scenario and the zero-shot settings yield similar results, showing that the 1000 dialogues on Cooking provide the same performance as 13000 dialogues on Wikipedia from QuAC\footnote{When randomly subsampling QuAC to the same size as DoQA the results on the cooking domain fall down to 36.5.}. The combination of both improves performance by 7 points ("Transfer" row), with small additional gains when adding Movies and Travel dialogues for further fine-tuning ("Transfer all" row). Note that the performance obtained for Cooking in the "Transfer" or "Transfer all" setting is \textbf{comparable to the one reported for QuAC}, where the training and test are from the same domain\footnote{BERT+HAE obtains 62.4 in QuAC \citep{qu2019}, 9 points higher than in DoQA Cooking, but note that QuAC contains more reference answers per question than DoQA, and thus the resulting F1 scores are higher. When evaluating BERT+HAE using a single reference answer in both datasets, the score is 45.9 on QuAC and 47.8 on the Cooking dataset of DoQA.}.  

Yet, the most interesting results are those for the \textbf{Travel and Movies} domains, which do not have access to in-domain training data on Travel or Movies. In this case, the native and \textbf{transfer results with no in-domain training} are as high as those for Cooking. These results show that it is not necessary to train for each domain in a FAQ, and that training data from other FAQ domains is highly reusable. 

The results obtained on out-of-domain test conversations (Movie and Travel) when trained on Wikipedia and Cooking are striking, as they are comparable to the in-domain results obtained for the Cooking test conversations. 
We hypothesize that when people write the answer documents in FAQ websites such as Stackexchange, they tend to use 
linguistic patterns that are common across domains such as Travel, Cooking or Movies. This is in contrast to Wikipedia text, which is produced with a different purpose, and might contain different linguistic patterns. As an example, in contrast to FAQ text, Wikipedia text does not contain first-person and second-person pronouns. We leave an analysis of this hypothesis for the future.

\begin{table}[t]
\centering
\begin{tabular}{llccccc} \hline
\multicolumn{2}{l}{Model}  & F1    & HEQ-Q  & F1-all\\ \hline
\multicolumn{2}{l}{Answer retrieval}&&&\\
&Top-1       & \textbf{37.2}    & \textbf{33.3} & \textbf{35.8}  \\ 
&Top-20:BERT       & 32.7 & 29.6   & 31.0   \\
&Top-20:BERT*IR & 36.1  &   32.9   &    34.4   \\
\hline
\multicolumn{2}{l}{Question retrieval}&&&\\
&Top-1       & \textbf{42.2}    & \textbf{36.76} & \textbf{41.1}  \\ 
&Top-20:BERT        & 35.8 & 31.2  & 34.3   \\
&Top-20:BERT*IR &  41.6  &   36.4 & 40.5   \\\hline
\end{tabular}
\caption{Results on the IR scenario (Cooking domain). See text for explanation} 
\label{tab:DoQA-ir}
\end{table}

Table \ref{tab:DoQA-ir} presents the results of the experiments on the \textbf{IR scenario}. The simplest Top-1 approach is the best performing for both question and answer retrieval strategies. We leave the exploration of more sophisticated techniques for future work. The results using question retrieval are very close to those in Table \ref{tab:transfer-results}. Given the large gap in the IR results in Section \ref{ssec:ir} for answer retrieval, it is a surprise to see a small 5 point decrease with respect to question retrieval. We found that there is a high correlation between the errors of the dialogue system and the answer retrieval system, which explains the smaller difference. In both retrieval strategies the \textbf{results are close} to the performance
obtaining when having access to the reference target passage.


\section{Conclusion and Future Work}
\label{sec:conclusion}

The goal of this work is to access the large body of domain-specific information in the form of Frequently Asked Question sites via conversational QA systems.  We have presented DoQA, a dataset for accessing Domain specific FAQs via conversational QA that contains 2,437 information-seeking dialogues on the Cooking, Travel and Movies domain (10,917 questions in total). These dialogues are created by crowdworkers that play the following two roles: the user asks questions about a certain topic posted in Stack Exchange, and the domain expert who replies to the questions by selecting a short span of text from the long textual reply in the original post. The expert can rephrase the selected span, in order to make it look more natural. In contrast to previous conversational QA datasets, our dataset responds to a real information need, is multi-domain, more natural and coherent. DoQA introduces a more realistic scenario where the passage with the answer needs to be retrieved.

Together with the dataset, we presented results of a strong conversational model, including transfer learning from Wikipedia QA datasets to our FAQ dataset. Our dataset and experiments show that it is possible to access domain-specific FAQs  using conversational QA systems with little or no in-domain training data, yielding quality which is comparable to those reported in QuAC. 

For the future, we would like to exploit the abstractive answers in our dataset, explore more sophisticated systems in both scenarios and perform user studies to study how real users interact with a conversational QA system when accessing FAQs.

\section*{Acknowledgments}

This research was partially supported by a Google Faculty Award, EU ERA-Net CHIST-ERA LIHLITH funded by the Agencia Estatal de Investigación (AEI, Spain) project PCIN-2017-118 and the Swiss National Science Foundation (SNF, Switzerland) project 20CH21 174237, project DeepReading (RTI2018-096846-BC21) supported by the Ministry of Science, Innovation and Universities of the Spanish Government, the Basque Government (DL4NLP KK-2019/00045 and excellence research group), project BigKnowledge (Ayudas Fundación BBVA a Equipos de Investigación Científica 2018) and  the NVIDIA GPU grant program. Jon Ander Campos enjoys a doctoral grant from the Spanish MECD.

\bibliography{acl2020}

\begin{thebibliography}{19}
\expandafter\ifx\csname natexlab\endcsname\relax\def\natexlab#1{#1}\fi

\bibitem[{Castelli et~al.(2019)Castelli, Chakravarti, Dana, Ferritto, Florian,
  Franz, Garg, Khandelwal, McCarley, McCawley, Nasr, Pan, Pendus, Pitrelli,
  Pujar, Roukos, Sakrajda, Sil, Uceda-Sosa, Ward, and
  Zhang}]{castelli2019techqa}
Vittorio Castelli, Rishav Chakravarti, Saswati Dana, Anthony Ferritto, Radu
  Florian, Martin Franz, Dinesh Garg, Dinesh Khandelwal, Scott McCarley, Mike
  McCawley, Mohamed Nasr, Lin Pan, Cezar Pendus, John Pitrelli, Saurabh Pujar,
  Salim Roukos, Andrzej Sakrajda, Avirup Sil, Rosario Uceda-Sosa, Todd Ward,
  and Rong Zhang. 2019.
\newblock \href {http://arxiv.org/abs/1911.02984} {{The TechQA Dataset}}.

\bibitem[{Chen et~al.(2017)Chen, Fisch, Weston, and Bordes}]{chen2017drqa}
Danqi Chen, Adam Fisch, Jason Weston, and Antoine Bordes. 2017.
\newblock \href {https://doi.org/10.18653/v1/P17-1171} {Reading {W}ikipedia to
  answer open-domain questions}.
\newblock In \emph{Proceedings of the 55th Annual Meeting of the Association
  for Computational Linguistics (Volume 1: Long Papers)}, pages 1870--1879,
  Vancouver, Canada. Association for Computational Linguistics.

\bibitem[{Choi et~al.(2018)Choi, He, Iyyer, Yatskar, Yih, Choi, Liang, and
  Zettlemoyer}]{choi2018quac}
Eunsol Choi, He~He, Mohit Iyyer, Mark Yatskar, Wen-tau Yih, Yejin Choi, Percy
  Liang, and Luke Zettlemoyer. 2018.
\newblock {QuAC: Question Answering in Context}.
\newblock In \emph{{Proceedings of the 2018 Conference on Empirical Methods in
  Natural Language Processing}}, pages 2174--2184.

\bibitem[{Devlin et~al.(2018)Devlin, Chang, Lee, and
  Toutanova}]{devlin2018bert}
Jacob Devlin, Ming-Wei Chang, Kenton Lee, and Kristina Toutanova. 2018.
\newblock {BERT: Pre-training of Deep Bidirectional Transformers for Language
  Understanding}.
\newblock \emph{arXiv preprint arXiv:1810.04805}.

\bibitem[{Dunn et~al.(2017)Dunn, Sagun, Higgins, G{\"{u}}ney, Cirik, and
  Cho}]{dunn2017searchqa}
Matthew Dunn, Levent Sagun, Mike Higgins, V.~Ugur G{\"{u}}ney, Volkan Cirik,
  and Kyunghyun Cho. 2017.
\newblock \href {http://arxiv.org/abs/1704.05179} {{SearchQA: {A} New Q{\&}A
  Dataset Augmented with Context from a Search Engine}}.
\newblock \emph{CoRR}, abs/1704.05179.

\bibitem[{He et~al.(2017)He, Balakrishnan, Eric, and Liang}]{he2017learning}
He~He, Anusha Balakrishnan, Mihail Eric, and Percy Liang. 2017.
\newblock Learning symmetric collaborative dialogue agents with dynamic
  knowledge graph embeddings.
\newblock In \emph{Proceedings of the 55th Annual Meeting of the Association
  for Computational Linguistics (Volume 1: Long Papers)}, pages 1766--1776.

\bibitem[{Iyyer et~al.(2017)Iyyer, tau Yih, and Chang}]{iyyer2017search}
Mohit Iyyer, Wen tau Yih, and Ming-Wei Chang. 2017.
\newblock \href {https://doi.org/10.18653/v1/P17-1167} {{Search-based Neural
  Structured Learning for Sequential Question Answering}}.
\newblock In \emph{Proceedings of the 55th Annual Meeting of the Association
  for Computational Linguistics (Volume 1: Long Papers)}, pages 1821--1831,
  Vancouver, Canada. Association for Computational Linguistics.

\bibitem[{Ko{\v{c}}isk{\'y} et~al.(2018)Ko{\v{c}}isk{\'y}, Schwarz, Blunsom,
  Dyer, Hermann, Melis, and Grefenstette}]{kocisky2018narrativeqa}
Tom{\'a}{\v{s}} Ko{\v{c}}isk{\'y}, Jonathan Schwarz, Phil Blunsom, Chris Dyer,
  Karl~Moritz Hermann, G{\'a}bor Melis, and Edward Grefenstette. 2018.
\newblock \href {https://doi.org/10.1162/tacl_a_00023} {The {N}arrative{QA}
  reading comprehension challenge}.
\newblock \emph{Transactions of the Association for Computational Linguistics},
  6:317--328.

\bibitem[{Nakov et~al.(2017)Nakov, Hoogeveen, M{\`a}rquez, Moschitti, Mubarak,
  Baldwin, and Verspoor}]{nakov2017semeval}
Preslav Nakov, Doris Hoogeveen, Llu{\'\i}s M{\`a}rquez, Alessandro Moschitti,
  Hamdy Mubarak, Timothy Baldwin, and Karin Verspoor. 2017.
\newblock Semeval-2017 task 3: Community question answering.
\newblock \emph{arXiv preprint arXiv:1912.00730}.

\bibitem[{Nakov et~al.(2016)Nakov, M{\`a}rquez, Moschitti, Magdy, Mubarak,
  Freihat, Glass, and Randeree}]{nakov2016semeval}
Preslav Nakov, Llu{\'\i}s M{\`a}rquez, Alessandro Moschitti, Walid Magdy, Hamdy
  Mubarak, Abed~Alhakim Freihat, Jim Glass, and Bilal Randeree. 2016.
\newblock \href {https://doi.org/10.18653/v1/S16-1083} {{S}em{E}val-2016 task
  3: Community question answering}.
\newblock In \emph{Proceedings of the 10th International Workshop on Semantic
  Evaluation ({S}em{E}val-2016)}, pages 525--545, San Diego, California.
  Association for Computational Linguistics.

\bibitem[{Nguyen et~al.(2016)Nguyen, Rosenberg, Song, Gao, Tiwary, Majumder,
  and Deng}]{nguyen2016ms}
Tri Nguyen, Mir Rosenberg, Xia Song, Jianfeng Gao, Saurabh Tiwary, Rangan
  Majumder, and Li~Deng. 2016.
\newblock {MS MARCO: A Human Generated MAchine Reading COmprehension Dataset}.
\newblock \emph{arXiv, abs/1611.09268}.

\bibitem[{Qu et~al.(2019)Qu, Yang, Qiu, Croft, Zhang, and Iyyer}]{qu2019}
Chen Qu, Liu Yang, Minghui Qiu, W.~Bruce Croft, Yongfeng Zhang, and Mohit
  Iyyer. 2019.
\newblock \href {http://arxiv.org/abs/1905.05412} {{BERT with History Answer
  Embedding for Conversational Question Answering}}.
\newblock \emph{CoRR}, abs/1905.05412.

\bibitem[{Rajpurkar et~al.(2018)Rajpurkar, Jia, and Liang}]{rajpurkar2018know}
Pranav Rajpurkar, Robin Jia, and Percy Liang. 2018.
\newblock {Know What You Don't Know: Unanswerable Questions for SQuAD}.
\newblock \emph{arXiv preprint arXiv:1806.03822}.

\bibitem[{Rajpurkar et~al.(2016)Rajpurkar, Zhang, Lopyrev, and
  Liang}]{rajpurkar2016squad}
Pranav Rajpurkar, Jian Zhang, Konstantin Lopyrev, and Percy Liang. 2016.
\newblock {SQuAD: 100,000+ questions for machine comprehension of text}.
\newblock \emph{arXiv preprint arXiv:1606.05250}.

\bibitem[{Reddy et~al.(2018)Reddy, Chen, and Manning}]{reddy2018coqa}
Siva Reddy, Danqi Chen, and Christopher~D Manning. 2018.
\newblock {CoQA: A conversational question answering challenge}.
\newblock \emph{arXiv preprint arXiv:1808.07042}.

\bibitem[{Talmor and Berant(2018)}]{talmor2018web}
Alon Talmor and Jonathan Berant. 2018.
\newblock \href {https://doi.org/10.18653/v1/N18-1059} {{The Web as a
  Knowledge-Base for Answering Complex Questions}}.
\newblock In \emph{Proceedings of the 2018 Conference of the North {A}merican
  Chapter of the Association for Computational Linguistics: Human Language
  Technologies, Volume 1 (Long Papers)}, pages 641--651, New Orleans,
  Louisiana. Association for Computational Linguistics.

\bibitem[{Trischler et~al.(2017)Trischler, Wang, Yuan, Harris, Sordoni,
  Bachman, and Suleman}]{trischler2017newsqa}
Adam Trischler, Tong Wang, Xingdi Yuan, Justin Harris, Alessandro Sordoni,
  Philip Bachman, and Kaheer Suleman. 2017.
\newblock \href {https://doi.org/10.18653/v1/W17-2623} {{N}ews{QA}: A machine
  comprehension dataset}.
\newblock In \emph{Proceedings of the 2nd Workshop on Representation Learning
  for {NLP}}, pages 191--200, Vancouver, Canada. Association for Computational
  Linguistics.

\bibitem[{Watanabe et~al.(2017)Watanabe, Dhingra, and
  Salakhutdinov}]{watanabeds2017questionanswering}
Yusuke Watanabe, Bhuwan Dhingra, and Ruslan Salakhutdinov. 2017.
\newblock \href {http://arxiv.org/abs/1703.08885} {{Question Answering from
  Unstructured Text by Retrieval and Comprehension}}.
\newblock \emph{CoRR}, abs/1703.08885.

\bibitem[{Yatskar(2018)}]{yatskar2018qualitative}
Mark Yatskar. 2018.
\newblock {A qualitative comparison of CoQA, SQuAD 2.0 and QuAC}.
\newblock \emph{arXiv preprint arXiv:1809.10735}.

\end{thebibliography}


\clearpage

\appendix

\section{FAQ Post Selection}
\label{app:postselect}
First, we downloaded the data dumps from September 2018 for cooking forum and September 2019 for travel and movies forums. We then removed threads with unaccepted answers. At this point we did a preliminary analysis of the cooking topic scores and the lengths of the answer passages. Regarding the scores, we realized that all topic scores were in the range $[-6, 240]$. After manually analysing some random samples, we concluded that even low scoring topics had a good quality, except for the ones with negative scores. Regarding the length of the answer passages, some of them were too long for our task (up to $2,960$ tokens), as very long passages makes the task very tedious. Taking all this into account, we applied the following filters to the topic-passage pairs for the three domains:

\begin{itemize}
    \item Topics with score $<=0$ are removed, as we are not interested in badly asked questions.
    \item Topic titles with more than one question mark are removed. The reason behind this filter is that we are interested in having the topic titles as the first question of our dialogues and we are not interested in having more than one question per dialogue turn. 
    \item The length of the answer passage has to be greater than $50$ and shorter than $250$ tokens. This way, we try to ensure that the answer passage is long enough for collecting dialogue, but not too long for avoiding tedious answer spotting.  
    \item Answers that contain HTML tags such as hyperlinks, images, code, etc. are removed.
\end{itemize}

\section{Amazon Mechanical Turk HIT Specifications}
\label{app:hit}
In order to select the workers in AMT, we defined the HIT with the following specifications:

\begin{itemize}
    \item HIT approval rate $\geq 98\%$.
    \item Approved HITs $\geq 1000$
    \item Location of the workers: English speaking countries. 
\end{itemize}    
    
We paid the workers $\$0.10$ for doing the HIT and a bonus of $\$0.33$ for each question or answer given during the task except for the ``I don't know sorry'' case where $\$0.05$ was paid. This difference in the payment motivates the workers to force themselves to find the actual answer in the passage, because answering ``I don't know'' is less demanding than searching for the correct answer span. The average price for each dialogue is  $\$3.2$. 

\section{Dialogue Collection Interfaces}
\label{app:interfaces}

\begin{figure*}[h!]
    \centering
    \includegraphics[width=0.9\textwidth]{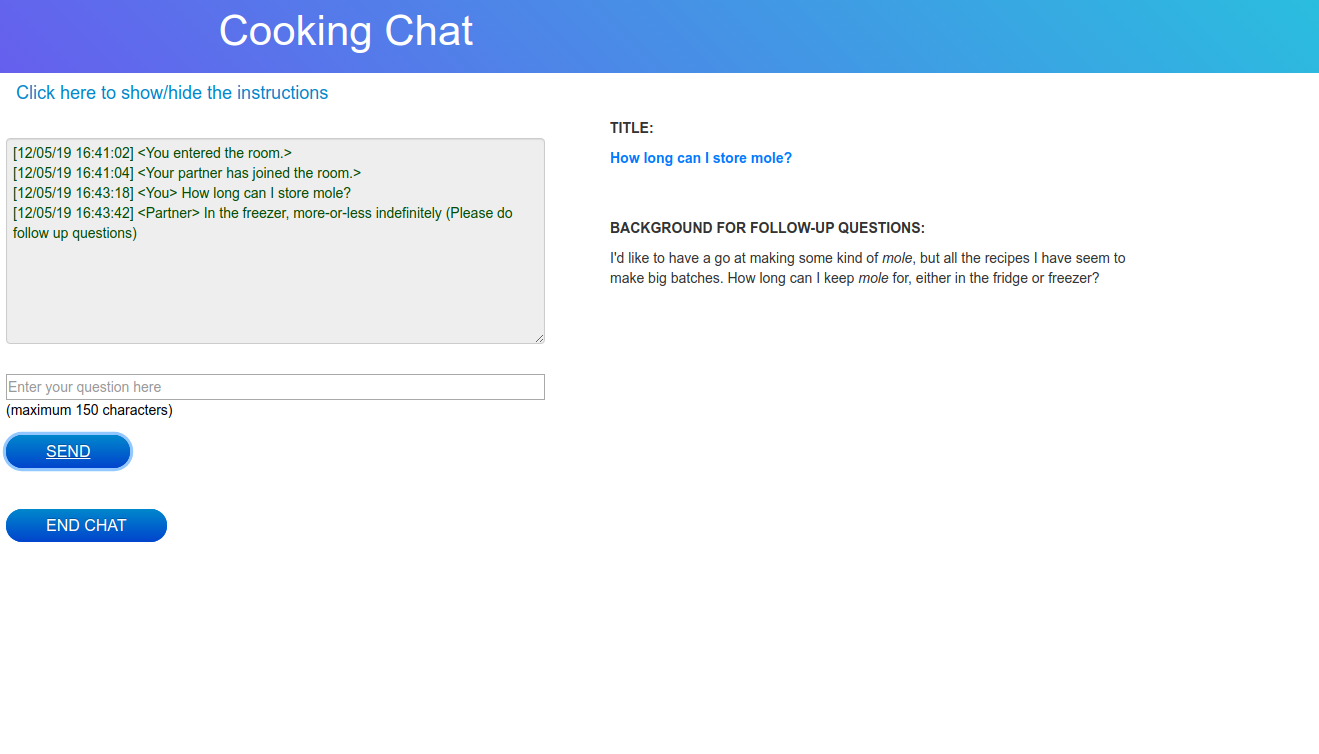}
    \caption{Dialogue collection interface for the user.}
    \label{fig:interf_user}
\end{figure*}

\begin{figure*}[h!]
    \centering
    \includegraphics[width=0.9\textwidth]{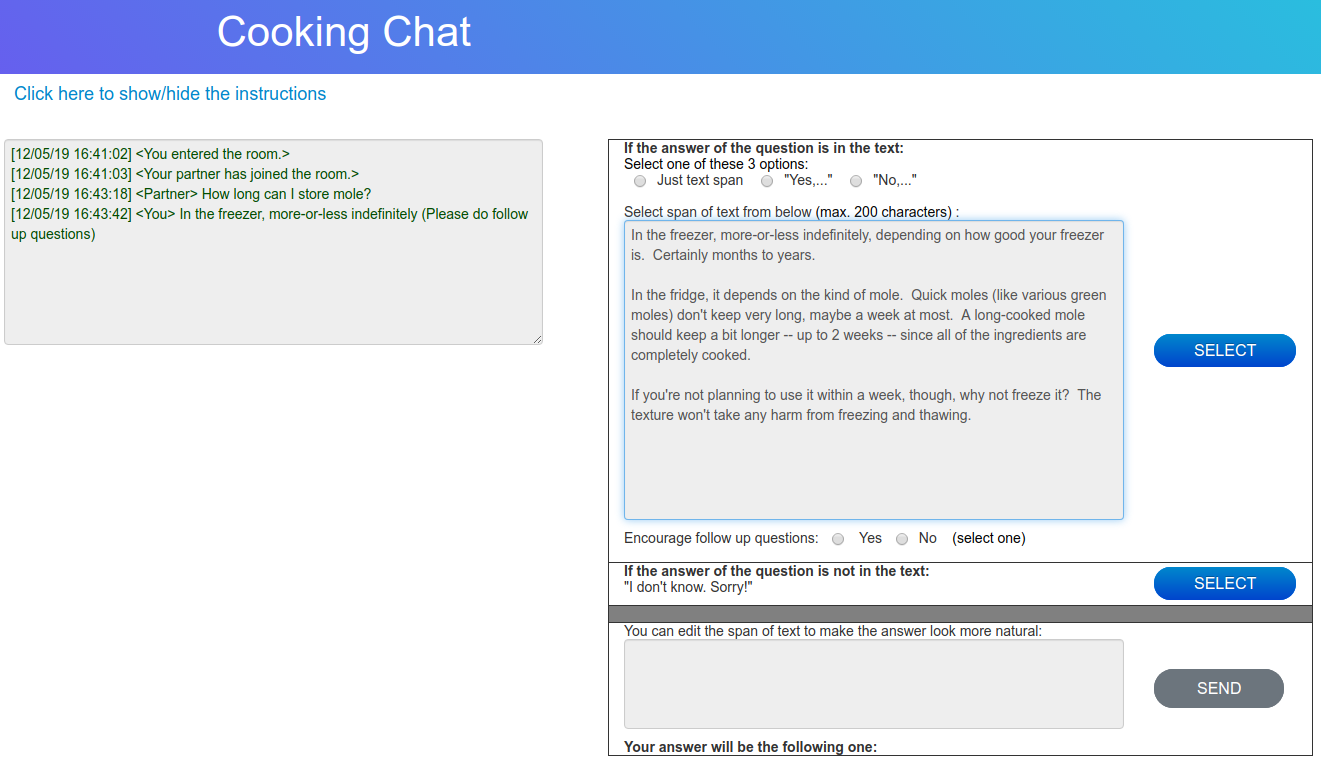}
    \caption{Dialogue collection interface for the expert.}
    \label{fig:interf_expert}
\end{figure*}

For dialogue collection, the worker carrying out the user role used the interface shown in Figure \ref{fig:interf_user} and one with the expert role used the interface displayed in Figure \ref{fig:interf_expert}.

\section{Multiple Answers Collection Interface}
\label{app:multiple}

\begin{figure*}[h!]
    \centering
    \includegraphics[width=0.9\textwidth]{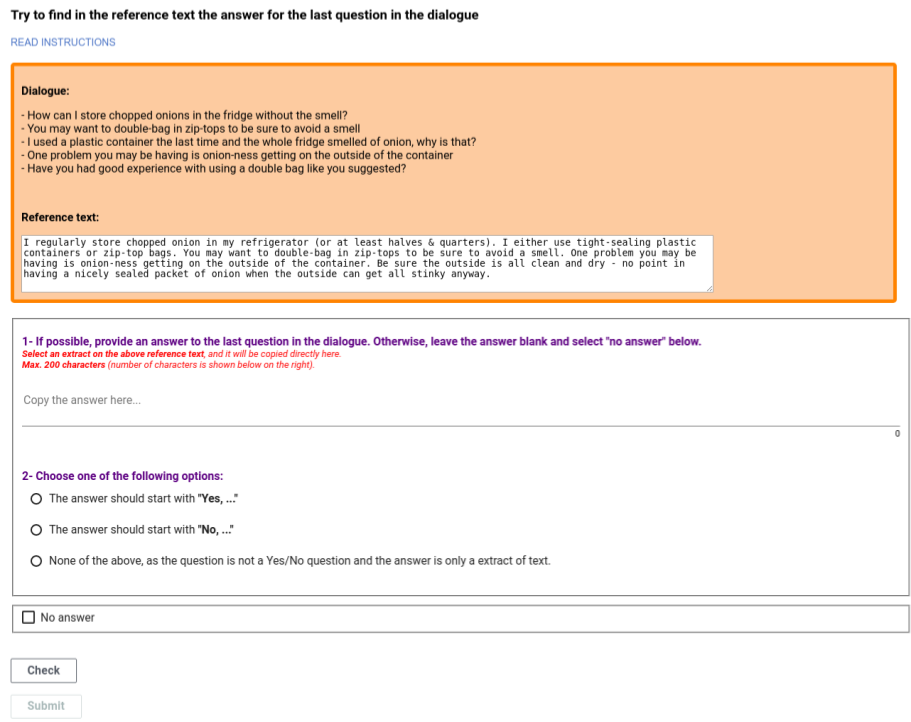}
    \caption{Multiple answers collection interface.}
    \label{fig:multiple_answers}
\end{figure*}

The interface used for multiple answers collection can be seen in Figure \ref{fig:multiple_answers}.

\end{document}